\documentclass[a4paper,conference]{IEEEtran}
\usepackage{cite}
\ifCLASSINFOpdf
  \usepackage[pdftex]{graphicx}
  \graphicspath{pictures}
\else
\fi
\ifCLASSOPTIONcompsoc
  \usepackage[caption=false,font=normalsize,labelfont=sf,textfont=sf]{subfig}
\else
  \usepackage[caption=false,font=footnotesize]{subfig}
\fi
\usepackage{hyperref}
\begin{document}
\title{4D-MultispectralNet: Multispectral Stereoscopic Disparity Estimation using Human Masks}

\author{\IEEEauthorblockN{Philippe Duplessis-Guindon}
\IEEEauthorblockA{LITIV lab.\\ Department of Computer and Software Engineering \\ Polytechnique Montreal \\
Email: philippe.duplessis-guindon@polymtl.ca}
\and
\IEEEauthorblockN{Guillaume-Alexandre Bilodeau}
\IEEEauthorblockA{LITIV lab.\\ Department of Computer and Software Engineering \\ Polytechnique Montreal \\
Email: gabilodeau@polymtl.ca}
}


%Telephone: (800) 555--1212, Fax: (888) 555--1212}


\maketitle

\begin{abstract}
Multispectral stereoscopy is an emerging field. A lot of work has been done in classical stereoscopy, but multispectral stereoscopy is not studied as frequently. This type of stereoscopy can be used in autonomous vehicles to complete the information given by RGB cameras. It helps to identify objects in the surroundings when the conditions are more difficult, such as in night scenes. This paper focuses on the RGB-LWIR spectrum. RGB-LWIR stereoscopy has the same challenges as classical stereoscopy, that is occlusions, textureless surfaces and repetitive patterns, plus specific ones related to the different modalities.  Finding matches between two spectrums adds another layer of complexity. Color, texture and shapes are more likely to vary from a spectrum to another. To address this additional challenge, this paper focuses on estimating the disparity of people present in a scene. Given the fact that people's shape is captured in both RGB and LWIR, we propose a novel method that uses segmentation masks of the human in both spectrum and than concatenate them to the original images before the first layer of a Siamese Network. This method helps to improve the accuracy, particularly within the one pixel error range.
\end{abstract}


% \fi
\IEEEpeerreviewmaketitle

\section{Introduction}

\begin{figure*}[!t]
\centering
\subfloat[RGB rectified image]{\includegraphics[width=1.75in]{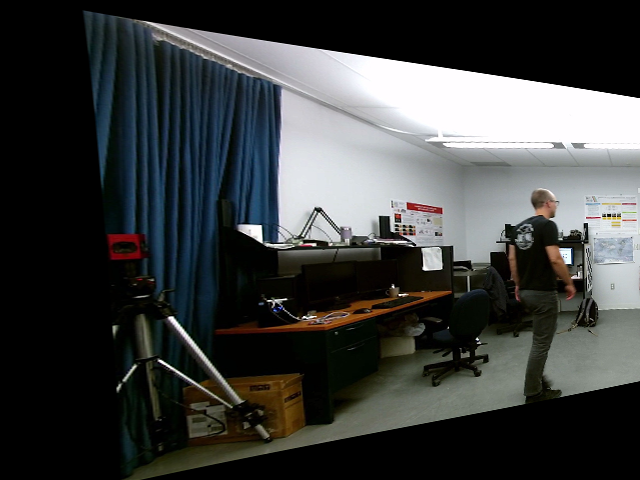}\label{rgb_original}}
\hfil
\subfloat[LWIR rectified image]{\includegraphics[width=1.75in]{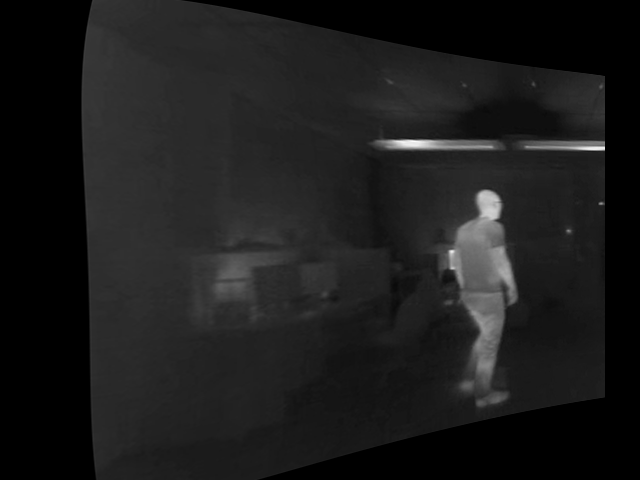}\label{lwir_original}}
\hfil
\subfloat[RGB mask]{\includegraphics[width=1.75in]{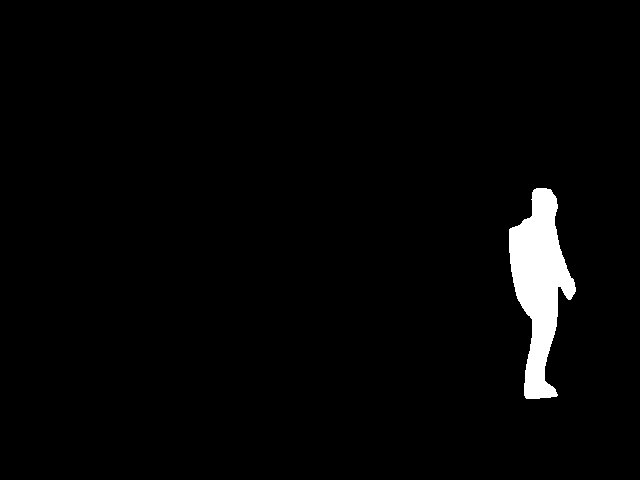}\label{rgb_mask}}
\hfil
\subfloat[LWIR mask]{\includegraphics[width=1.75in]{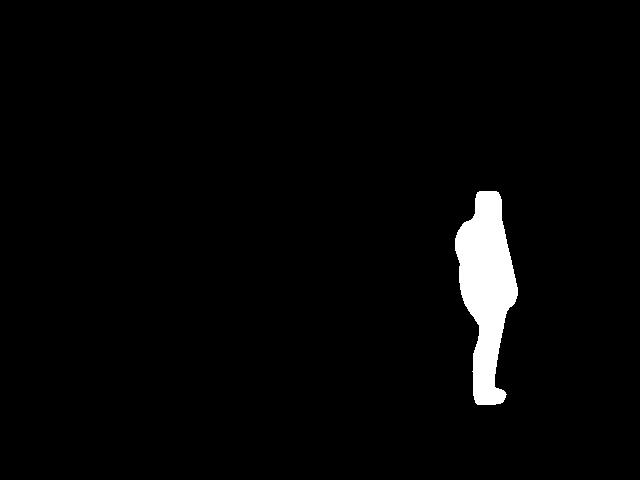}\label{lwir_mask}}
\hfil
\subfloat[RGB 4-channel image]{\includegraphics[width=1.75in]{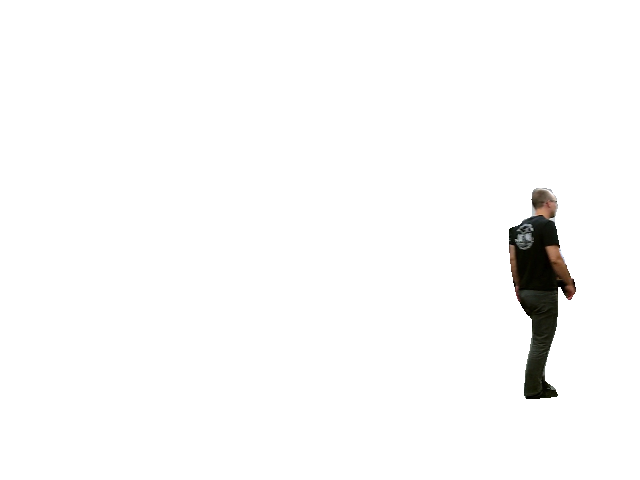}\label{rgb_patern_ex}}
\hfil
\subfloat[LWIR 4-channel image]{\includegraphics[width=1.75in]{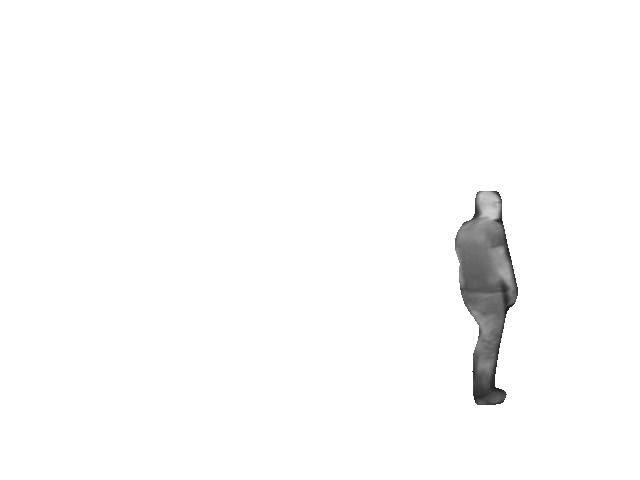}\label{lwir_patern_ex}}
\caption{RGB and LWIR image process from the dataset}
\label{patern_ex}
\end{figure*}

This paper focuses on estimating the pixel disparities between an RGB image and a long wave infrared (LWIR) image, commonly called a thermal image. Those disparities can be used to align the visible and thermal images to generate an augmented image. Adding a new imaging spectrum helps improving the accuracy in some complex situations (e.g. low light, fog, smoke) for tasks like object detection and tracking. This is very useful when a scene is dark, or when there is some low contrast between the person and the background.  In multispectral stereo, we cannot simply rely on pixel patterns like we do in classical stereo. We can see an example of this in Fig. \ref{patern_ex}. The shirt logo on the RGB (Fig. \ref{rgb_original}) image does not appear on the LWIR (Fig. \ref{lwir_original}) image. In classical stereo, the logo on the shirt would have been a great way to estimate the disparity. We can also observe that the heat emitted through the shirt is not evenly distributed on the surface of the shirt. So in the LWIR image, the shirt region has some color variation, and on the RGB image, the color is uniformly black on the shirt surface. This is one of the difficulties that makes RGB-LWIR stereoscopy very challenging. 

To address this challenge, we are focusing on humans because they are the important subject in several tasks. Furthermore, in several situations, human silhouettes can be captured in both RGB and thermal images, as the human silhouette in a LWIR images is coming from the heat emitted from the body. \emph{Our goal is to design a method that estimates the disparity between the pixels of two human silhouettes. We propose as a novel idea to integrate the pixel masks of people in a neural network based on a  Siamese architecture.} Our proposed network, 4D-MultispectralNet, concatenates the masks to the original images, and then passes them through a Siamese network. This allows the network to attend specifically to pixels included in the mask and to account better for object boundaries where there are disparity discontinuities. The proposed method is inspired by Beaupré et al. \cite{beaupre2020domain}, where as oppose to RGB stereo that uses siamese networks, the main idea was not to share the weights between the two CNNs, since each spectrum has its own characteristics. In this paper we build upon this idea.  Mask R-CNN \cite{Maskrcnn}, using detectron2 \cite{wu2019detectron2} is applied on every image to create a segmentation mask of the humans in the scene. The segmentation masks are applied on both RGB and LWIR images. This steps provides more information about the positions of the humans in the scene as well as on their boundaries where there are potential disparity discontinuities. We show that this improves the disparity accuracy. With our proposed method, we achieve better accuracy on the less than 1 pixel error metric.

\section{Literature review}
\subsection{RGB Stereo}
Many papers studied RGB stereoscopy. The first paper that used a Convolutional Neural Networks (CNN) to solve this problem was Zbontar and LeCun \cite{7298767}. Their method was to take a $9\times 9$ region on the left and on the right image. The goal was to learn the similarity between the two regions, and since the location of each region was known, the disparity could be established. For example, if the method identified that the region at the position $(x_1, y)$ corresponded to the region at position $(x_2, y)$, the disparity between those two patches was $|x_2 - x_1|$. The architecture of their CNN consists of a convolution layer followed by fully connected layer for each $9\times 9$ images. Afterwards, the output of these two subnetworks are combined and passed through fully connected layers. The output is a binary classification that determines if the two inputs are the same.

Another pioneering work is by Luo et al. \cite{efficientDL}. Their approach consists of taking on the left image, a $9\times 9$ image patch, and on the right image, a patch of the same height, but with the full width of the image. A feature vector is later created for the $9\times 9$ patch and a feature volume is created for the wider right patch. Afterwards, a correlation product is made between the feature vector of the left image (the $9\times 9$ patch) and every vector in the feature volume of the right image. These correlation products give a probability distribution of finding the current disparity for every vector. This correlation vector to merge the left and right feature vectors makes the stereo more efficient.

Kendall et al. \cite{kendall2017endtoend} introduced the GC-Net, which was the first work to propose an end-to-end architecture. They used a Siamese network to extract the features, and then they used a 3D convolution to learn from these feature vectors. A regression is made afterwards to give the resulting disparity maps. Many other methods built over these last three papers, with various variations and improvements. For example, \cite{chang2018pyramid} used a Spatial Pyramid Pooling Module to extract the important features in the feature vectors, and they also used an hourglass network for the cost volume regularization and disparity regression. 

\subsection{Multispectral Stereo}

Before, the neural network era, multispectral stereoscopy were based on matching feature points. One way to do so was to use SIFT \cite{Lowe2004} as a feature descriptor. From  SIFT emerged MSIFT \cite{5995637}. MSIFT adapted SIFT for multispectral purposes by improving the correlation between the RGB channel for a RGB-NIR pair of images. Instead of using feature descriptors, some methods are using window-based method to find the matches of both images, such mutual information \cite{466930}, HOG \cite{HOG}, SSD \cite{LITIV2014}, LSS \cite{LSS} and more. According to Bilodeau et al. \cite{LITIV2014},  Mutual information is the more accurate method among the window-based methods.

If we focus on more recent work studying RGB/LWIR stereo, Beaupré et al. \cite{Beaupre_2019_CVPR_Workshops} proposed a method that consists of having two Siamese networks. They share parameters. One network computes the disparity from RGB to LWIR and the other the disparity from LWIR to RGB. The principle of each siamese network is similar to the method used by Luo et al \cite{efficientDL}. The first part of the network compares a small patch of an RGB image with a patch in the LWIR image of the same height, but of the full width of the original image. The RGB patch is a small patch around the disparity point that is to be estimated, and it is fed in the network as the first input. The other input is the wider LWIR image. A correlation is done with every translation possible to find the corresponding disparity. For the other Siamese network, the same principle applies, but the small patch is the LWIR image at the given disparity and the wider image is the RGB image. To
select the final disparity a summation layer is in place to sum up the prediction vector from each branch. The final disparity will be the max element in this vector.

A following work by Beaupré et al. \cite{beaupre2020domain} modified the previous approach. The main difference between this work and the previous is that the weights are not shared between CNNs in this architecture. This significantly improved the results. The reason why the weights should not shared is because of the nature of each spectral image. In most cases when Siamese network are used, the inputs are similar in term of color, shape and contrast. This is not the case for RGB-LWIR stereo matching. Both images do not have a lot in common. The only similarity between images is mostly the shape of the objects. The shape is not even exactly the same since the IR images are capturing the body's heat, and it makes the edges less accurate in comparison to RGB images. This work inspired our approach.

RGB and LWIR is not the only type of multispectral stereo studied. Several works are doing RGB-NIR Stereo. Aguilera et al. \cite{Aguilera2016} made a study which compares three different CNNs architectures to see if they are more efficient than classical method mentioned earlier. A following work by Aguilera et al. \cite{Aguilera2017} introduced the quadruplet networks. This network takes two matching pairs of images. This helps because the network has two pairs of positive example and four pairs of negative example. Same as RGB/LWIR, there is not a lot of dataset for RGB/NIR stereo. Zhi et al. \cite{Zhi_2018_CVPR} created a dataset to overcome this problem. With this dataset they are transforming an RGB image into the NIR spectrum. They are using this newly generated image to do self-supervision.

\begin{figure*}[h]
\vskip 0.2in
\begin{center}
\centerline{\includegraphics[width=5in]{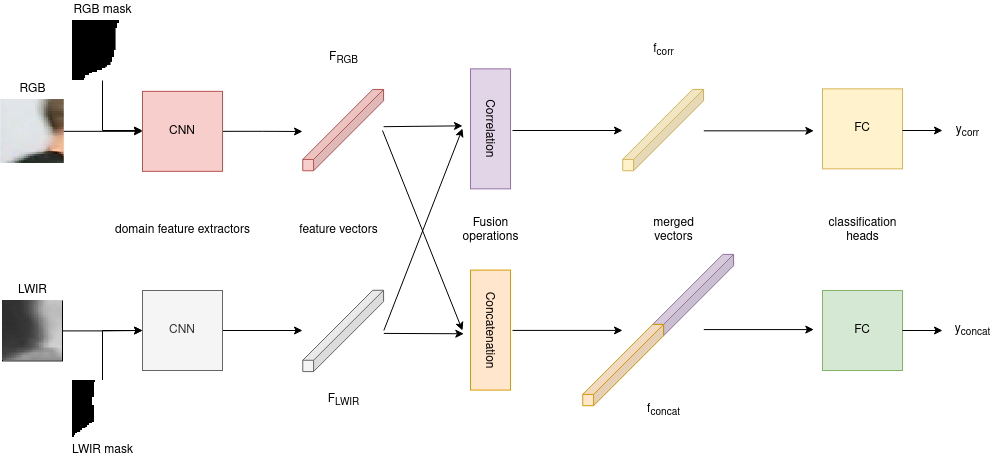}}
\caption{Our architecture Siamese CNNs architecture}
\label{sCNNs}
\end{center}
\vskip -0.2in
\end{figure*}

\section{Proposed method}
\subsection{Overview}

Our method is inspired from the works mentioned previously and uses mask R-CNN, using detectron2 \cite{wu2019detectron2} on every image to create a segmentation mask on the humans in the scene. The segmentation masks are applied on both RGB and LWIR images. This step provides more information about the positions of the humans in the scene and also about the boundaries of the human where disparity discontinuities occur. By doing so, our method is focusing only on human disparity estimation, but in most applications, humans are the subject of interest. Note that our method would work also with warm-blooded animals. The masks are merged with the RGB and LWIR images. The data will now consists in 4-channel images, the fourth channel being the generated mask by detectron2 \cite{wu2019detectron2}.

\subsection{Using object masks for disparity estimation}

This section is the core of our method. We noticed that previous methods had difficulty obtaining reliable disparities near object boundaries. This motivated us to include object masks in our stereo method. Now, the question was how to include this information in a CNN. We found that the best approach was to let the network learn how the use the mask information by concatenating the masks with the input images.  The first step of this method was to generate every mask with Detectron2 \cite{wu2019detectron2}. We used a pretrained version on the COCO dataset \cite{lin2015microsoft}. We considered only the "person" class to obtain the human silhouettes. We can see the outputs of Detectron2 on RGB and LWIR images in Fig. \ref{rgb_mask} and Fig. \ref{lwir_mask}, respectively. Although Detectron2 was not trained with LWIR, the results are good enough for our purpose. With these masks, we created 4-channel images that will be passed in our Siamese neural network. Fig. \ref{rgb_patern_ex} and Fig. \ref{lwir_patern_ex} are a representation of the final image that goes through the network. It is important to note that Fig. \ref{rgb_patern_ex} and Fig. \ref{lwir_patern_ex} are only a visual representation of the mask concatenation. The white pixel do not represent a RGB value of (255, 255, 255) but represent a pixel with a fourth layer valued at 255. A way to visualize this is that the mask represents the alpha of a RGBA image. The value of a white pixel in Fig. \ref{rgb_patern_ex} or Fig. \ref{lwir_patern_ex} is truly $(R_{i,j}, G_{i,j}, B_{i,j}, 255)$. 

            \begin{table}[h]
            \centering
            \caption{Our proposed network architecture.}
            \begin{tabular}{|lcc|}
            \hline
            \multicolumn{1}{|c|}{Name}   & \multicolumn{1}{c|}{Layer structure} & Output dimension \\ \hline
            \multicolumn{1}{|l|}{Input}  & \multicolumn{1}{l|}{}                & 36 x 36 x 4      \\ \hline
            \multicolumn{3}{|c|}{CNN}                                                              \\ \hline
            \multicolumn{1}{|l|}{conv 1} & \multicolumn{1}{c|}{5 x 5, 32}       & 32 x 32 x 32     \\ \hline
            \multicolumn{1}{|l|}{conv 2} & \multicolumn{1}{c|}{5 x 5, 64}       & 28 x 28 x 64     \\ \hline
            \multicolumn{1}{|l|}{conv 3} & \multicolumn{1}{c|}{5 x 5, 64}       & 24 x 24 x 64     \\ \hline
            \multicolumn{1}{|l|}{conv 4} & \multicolumn{1}{c|}{5 x 5, 64}       & 20 x 20 x 64     \\ \hline
            \multicolumn{1}{|l|}{conv 5} & \multicolumn{1}{c|}{5 x 5, 128}      & 16 x 16 x 128    \\ \hline
            \multicolumn{1}{|l|}{conv 6} & \multicolumn{1}{c|}{5 x 5, 128}      & 12 x 12 x 128    \\ \hline
            \multicolumn{1}{|l|}{conv 7} & \multicolumn{1}{c|}{5 x 5, 256}      & 8 x 8 x 256      \\ \hline
            \multicolumn{1}{|l|}{conv 8} & \multicolumn{1}{c|}{5 x 5, 256}      & 4 x 4 x 256      \\ \hline
            \multicolumn{1}{|l|}{conv 9} & \multicolumn{1}{c|}{4 x 4, 256}      & 1 x 1 x 256      \\ \hline
            \multicolumn{3}{|c|}{FC}                                                               \\ \hline
            \multicolumn{1}{|l|}{fc 1}   & \multicolumn{1}{c|}{256/512, 128}    & 1 x 128          \\ \hline
            \multicolumn{1}{|l|}{fc 2}   & \multicolumn{1}{c|}{128, 64}         & 1 x 64           \\ \hline
            \multicolumn{1}{|l|}{fc 3}   & \multicolumn{1}{c|}{64, 2}           & 1 x 2            \\ \hline
            \end{tabular}
            
            \label{layers_architecture}
            \end{table}
\subsection{Network architecture}

Our architecture is inspired by the methods of  Beaupré et al. \cite{Beaupre_2019_CVPR_Workshops, beaupre2020domain}. We have adapted their architecture to add a fourth channel in the input of the Siamese networks for our segmentation masks. Fig. \ref{sCNNs} illustrates our architecture. Our Network takes as input one RGB patch and one LWIR patch to extract their features. Both patches are $36\times 36$ as it allows to capture a reasonably large local neighborhood around each disparity point to evaluate.

Since each patch is coming from different spectrum, two separated feature extractor were shown to give better results \cite{beaupre2020domain}. So each patch goes through its own CNN. Both CNNs have the same layer architecture which is summarized in Table \ref{layers_architecture}. Even if both CNNs have the same layers, they do not share weights. This is because both spectrum do not contain the same information. Having two distinct feature extractor helps extracting specific details from each spectrum. Each CNN outputs a $256D$ feature vector, represented by $F_{RGB}$ and $F_{LWIR}$ in Fig. \ref{sCNNs}.

    We apply two fusion operations on those feature vectors. A correlation and a concatenation \cite{guo2019groupwise}. Both have pros and cons. The correlation fusion is faster and more memory efficient, but some features from both spectrum are lost during the operation. For the concatenation operation, no features are lost but there is a trade of with the compute time and the memory space. The correlation operation outputs a $256D$ feature vector, represented by $f_{corr}$ on Fig. \ref{sCNNs}. The concatenation operation outputs a $512D$ feature vector which is represented by $f_{concat}$ in Fig. \ref{sCNNs}. Each correlation and concatenation vectors are going through a fully connected network. One for the concatenation branch and the other one is for the correlation branch. Once again weights are not shared between each fully connected networks. Each branch outputs a 2D $y$ vector that represent a classification vector. Theses are represented by $y_{corr}$ and $y_{concat}$ on Figure \ref{sCNNs}.

\subsection{Training}

For training, we take patches around the ground truth in each spectrum in rectified images. We refer to these patches around the points as $P_{RGB}$ and $P_{LWIR}$.

Similarly to Beaupré et al. \cite{beaupre2020domain} work, we are using two cross-entropy loss functions for both branch in our network, one for the correlation and another one for the concatenation branch. They are given by
\begin{equation}
    loss_{corr/concat} = -1/N \sum_{i=1}^N gt_i log(y_i),
\end{equation}

\noindent where $N$ represent the number of data points, $gt_i$ the ground truth, which is 0 or 1 if the patches are the same, and $y_i$ is the similarity probability, which is represented by $y_{corr}$ or $y_{concat}$.

The total loss function is given by the sum of both loss in both branches by

\begin{equation}
    loss_{total} = loss_{corr} + loss_{concat}.
\end{equation}

\subsection{Disparity estimation}

To evaluate the disparity, we have to establish a max disparity, refer here as $d_{max}$. Next, we are adding half of this distance on both side of the LWIR patch, making it wider. However, the RGB patch keeps the same size. After passing these patches, $F_{RGB}$ will be a 256D vector and $F_{LWIR}$ will be a tensor of $256 \times d_{max}$. We then pass every feature vector of the tensor in the fusions operation with the unchanged $F_{RGB}$. Both vectors are then forwarded in the fully connected layers as explained earlier. Each $y_{corr}$ and $y_{concat}$ are now corresponding to the probability of the patches being the same or not. 
So for every LWIR patch we have a matching probability that this patch is corresponding to the RGB patch. The disparity of this current branch is then the index with the highest probability. This is given by

\begin{equation}
    \hat{d}_{corr/concat} = argmax(d),
\end{equation}

\noindent where $d$ is the disparity output vector of size $d_{max}$ of each branch ($y_{corr}$ and $y_{concat}$).

The final disparity is an average of the best disparity from each branch $\hat{d}_{corr}$ and $\hat{d}_{concat}$ and is given by
\begin{equation}
    \hat{d} = \frac{\hat{d}_{corr} + \hat{d}_{concat}}{2}.
\end{equation}

\section{Experiments}

The code of 4D-MultispectralNet is available on GitHub at the following link: \url{https://github.com/philippeDG/4D-MultispectralNet}     

\subsection{Dataset and metrics}

We used two datasets: the LITIV 2014 \cite{LITIV2014} dataset and the LITIV 2018 \cite{LITIV2018}. Our method was evaluated with cross-validation and trained/tested using different folds, mixing both datasets for training, validation and testing. We used the same folds as Beaupré et al. \cite{beaupre2020domain}. The datasets feature several actors moving in a room.

It is to note that few files are missing from the original datasets. Therefore, for a fair comparison, we re-run the Beaupré et al. \cite{beaupre2020domain} method on the slightly incomplete dataset. We were not able to reproduce the exact same results as their paper for this reason. 

We have evaluated our method with the recall metric as previous works. The following equation represents the performance measure used in this paper:
    \begin{equation}
        Recall = \frac{1}{N}\sum_{i=1}^N |\hat{d}_i - gt_i| \leq n
        \label{recal_formula}
    \end{equation}

In this equation $N$ stands for the number of points to be evaluated, $\hat{d}_i$ represent the evaluated disparity at a given point, $gt_i$ is the ground truth at the same given point, and lastly $n$ represent the allowed correspondence error in pixels.

           \begin{table*}[h]
             \caption{Results on LITIV 2014 compared to SOTA Methods. Results are the mean of the 3 videos. \dag We re-run their code with the slightly incomplete dataset. \ddag: results on the complete dataset. \textbf{Boldface: best results}.}
            \centering
            \begin{tabular}{|c||c|c|c|}
            \hline
            Method                                                                       & $\leq$ $1$ $pixel$ $error$ & $\leq$ $3$ $pixel$ $error$ & $\leq$ $5$ $pixel$ $error$        \\ \hline
            4D-MultispectralNet                                                                  & \textbf{57.52 $\pm$ 2.32}  & 88.74 $\pm$ 0.99           & \textbf{98,55 $\pm$ 0.39}         \\ \hline
            Domain Siamese CNN \cite{beaupre2020domain} \dag            & 56.25 $\pm$ 3.47           & \textbf{89.95 $\pm$ 0.39}  & 98.53 $\pm$ 0.43                  \\ \hline
            Siamese CNN \ddag \cite{Beaupre_2019_CVPR_Workshops}                               & -                          & $77.9 \pm 5.04$            & -                                 \\ \hline
            St-charles \cite{LITIV2018} \ddag                       & 48.2   $\pm$  3.95                    & -                       & -                                 \\ \hline
            Mutual Information \cite{LITIV2014} ($40 \times 130$) \ddag                       & -                          & 83.3                       & -                                 \\ \hline
            Mutual Information \cite{LITIV2014} ($20 \times 130$)\ddag                        & -                          & 77.5                       & -                                 \\ \hline
            Mutual Information \cite{LITIV2014} ($10 \times 130$) \ddag                       & -                          & 64.9                       & -                                 \\ \hline
            Fast Retina Keypoint \cite{LITIV2014}($40 \times 130$)   \ddag                    & -                          & 64.1                       & -                                 \\ \hline
            Local Self-Similarity \cite{LITIV2014, LITIV2018}($40 \times 130$)\ddag                      & 22.6 $\pm$ 10.66                          & 73.4                       & -                                 \\ \hline
            Sum of Squared Difference \cite{LITIV2014}($40 \times 130$) \ddag                 & -                          & 65.6                       & -                                 \\ \hline
            \end{tabular}
            \label{method_comp_2014}
            \end{table*}

\begin{table}[h]
             \caption{Results on LITIV 2018 compared to SOTA Methods. Results are the mean of the 3 videos. \dag We re-run their code with the slightly incomplete dataset. \ddag: results on the complete dataset. \textbf{Boldface: best results}.}
            \centering
            \begin{tabular}{|c|c|}
            \hline
            Method                                                                       & $\leq$ $1$ $pixel$ $error$      \\ \hline
            4D-MultispectralNet                                                                   & \textbf{60.45 $\pm$ 4.38}                 \\ \hline
            Domain Siamese CNN \cite{beaupre2020domain} \dag                             & 44.40 $\pm$ 3.70                  \\ \hline
            St-charles \cite{LITIV2018} \ddag                                            & 42.23   $\pm$  16.27               \\ \hline
            \end{tabular}
           % \begin{tablenotes}
             %   \item[\textdagger] 
            %\end{tablenotes}

            \label{method_comp_2018}
            \end{table}

            \begin{table}[h]
                     \caption{Results on LITIV 2014 for the 1 pixel error.}
            \centering
            \begin{tabular}{|l||l|l|}
            \hline
            \multicolumn{1}{|c||}{} & \multicolumn{1}{c|}{Beaupré et al. \cite{beaupre2020domain}} & Our Method \\ \hline
            Fold 1                 & 57.38                        & \textbf{57.91}      \\ \hline
            Fold 2                 & 52.35                        & \textbf{55.02}      \\ \hline
            Fold 3                 & 59.02                        & \textbf{59.63}   \\ \hline
            \end{tabular}
   
            \label{method_fold_comp_2014}
            \end{table}
            
            \begin{table}[h]
                     \caption{Results on LITIV 2018 for the 1 pixel error.}
            \centering
            \begin{tabular}{|l||l|l|}
            \hline
            \multicolumn{1}{|c||}{} & \multicolumn{1}{c|}{Beaupré et al. \cite{beaupre2020domain}} & Our Method \\ \hline
            Fold 1                 & 48.0                         & \textbf{64.23}    \\ \hline
            Fold 2                 & 44.6                         & \textbf{55.64}     \\ \hline
            Fold 3                 & 40.6                         & \textbf{61.49}      \\ \hline
            \end{tabular}
   
            \label{method_fold_comp_2018}
            \end{table}

\subsection{Data processing and data augmentation}

For data augmentation, we used the same method as Beaupré at al. \cite{beaupre2020domain}. The datasets are not very big, so data augmentation is needed for good performance. The first data augmentation technique consists of making the surroundings of a ground truth as if it was a ground truth itself. That is called cross-duplication \cite{beaupre2020domain}. So for a pixel  with a Manhattan distance of one we consider that they all have the same disparity. This making the dataset 5 times bigger. Another technique used to make more data is the mirroring over the $y$ axis. This additionally doubles the number of data points. We also generated every masks of the rectified images with Detectron2 \cite{wu2019detectron2}. 

\subsection{Comparison with state-of-the-art methods}

Table \ref{method_comp_2014} gives the results of our proposed method, 4D-MultispectralNet, compared to several state-of-the-art (SOTA) methods on LITIV 2014. This table was obtained with the mean of three folds. First, we can note that results for Beaupré et al. are slightly lower for the $n=3$ precision, but are higher for the $n=1$ and $n=5$ precision compared to their original paper since the dataset is not exactly the same. From table \ref{method_comp_2014}, we can see that our method have similar results for the $n = 3$ precision, but for $n = 1$ and $n = 5$ precision, the recall score improves. By comparing the results in table \ref{method_comp_2014} we can see that our results are improving in the $\leq$ $1$ $pixel$ $error$ ($n=1$) and for $\leq$ $5$ $pixel$ $error$ ($n=5$) compared to Domain Siamese CNN. This shows the benefit of using masks.

Table \ref{method_comp_2018} shows the results on the LITIV 2018 dataset \cite{LITIV2018}. This table also was obtained with the mean of three folds. Since other methods did not test on $\leq$ $3$ $pixel$ $error$ ($n=3$) or $\leq$ $5$ $pixel$ $error$ ($n=5$) on the LITIV 2018 dataset, \cite{LITIV2018} we are only showing the results on the $\leq$ $1$ $pixel$ $error$ ($n=1$). We can see a significant improvement on the recall score, and the standard deviation indicates that every fold with our method is better than previous works.  We can see that our method really improve the recall score thanks again to the use of the person masks.

Finally, if we detail the results of the folds for both datasets,  we can see in the table \ref{method_fold_comp_2014} that our method improved over Domain Siamese CNN that use an architecture similar to ours, but without masks, for the three folds of LITIV 2014. This is confirmed in Table \ref{method_fold_comp_2018}, where we can see that our method really improved the method for the folds on the LITIV 2018 dataset \cite{LITIV2018}.

\begin{table*}[h]
            \caption{Ablation study}
            \centering
            \begin{tabular}{|l||ccc||ccc||ccc|}
            \hline
                   & \multicolumn{3}{c||}{4d correlation branch}                      & \multicolumn{3}{c||}{4d concatenation branch}                    & \multicolumn{3}{c|}{4D-MultispectralNet}                              \\ \hline
                   & \multicolumn{1}{c|}{n1}    & \multicolumn{1}{c|}{n3}    & n5    & \multicolumn{1}{c|}{n1}    & \multicolumn{1}{c|}{n3}    & n5    & \multicolumn{1}{c|}{n1}    & \multicolumn{1}{c|}{n3}    & n5    \\ \hline
            Fold 1 & \multicolumn{1}{c|}{53.33} & \multicolumn{1}{c|}{86.72} & 97.70 & \multicolumn{1}{c|}{45.95} & \multicolumn{1}{c|}{82.31} & \textbf{99.14} & \multicolumn{1}{c|}{\textbf{57.91}} & \multicolumn{1}{c|}{\textbf{89.51}} & 98.31 \\ \hline
            Fold 2 & \multicolumn{1}{c|}{\textbf{62.47}} & \multicolumn{1}{c|}{\textbf{93.05}} & \textbf{99.46} & \multicolumn{1}{c|}{49.42} & \multicolumn{1}{c|}{83.10} & 96.93 & \multicolumn{1}{c|}{55.02} & \multicolumn{1}{c|}{87.61} & 98.35 \\ \hline
            Fold 3 & \multicolumn{1}{c|}{58.31} & \multicolumn{1}{c|}{86.68} & 97.77 & \multicolumn{1}{c|}{56.82} & \multicolumn{1}{c|}{84.76} & 98.00 & \multicolumn{1}{c|}{\textbf{59.63}} & \multicolumn{1}{c|}{\textbf{89.1}}  & \textbf{99.0}  \\ \hline
            \end{tabular}
            
            \label{ablation_study}
            \end{table*}

\subsection{Ablation study}

Table \ref{ablation_study} shows the ablation study made on both of the correlation and concatenation branches. So both CNNs extracted features from each patch, but only one fusion operation at a time was done.  For the first and third fold we can see that the combination of both branches gives better results than each branch separately. Although, for the second fold, the correlation branch gives better result than the concatenation branch and both branches combined. Given theses observations, it is better as a general rule to keep both branches in the architecture.

Results between our methods and Beaupre et al. \cite{beaupre2020domain} are directly showing the effect of the masks for our problem. Tables \ref{method_comp_2014} and \ref{method_comp_2018} are showing the beneficial impact of the masks on every fold because it helps in locating person boundaries.

\section{Conclusion}
The present work shows an efficient method that concatenates a segmentation mask on the initial data to improves the accuracy of disparity estimation. With the mask concatenation we achieved better results than other SOTA work in RGB-LWIR stereo matching on the 1 pixel error level of precision. The $\leq$ $1$ $pixel$ $error$ is better because of the masks that allow to local object boundaries. Using masks is not detrimental since we are keeping a similar accuracy as $\leq$ $3$ $pixel$ $error$ and $\leq$ $5$ $pixel$ $error$.

\section*{Acknowledgment}

This work was supported by the National Sciences and Engineering Research Council of Canada (NSERC).


%\newpage
% http://mirror.ctan.org/biblio/bibtex/contrib/doc/z
%\bibliographystyle{IEEEtran}
\nocite{*}
\bibliographystyle{IEEEtran}
\bibliography{IEEEabrv,ref.bib}

\end{document}